\documentclass[conference]{IEEEtran}
\usepackage{cite}
\usepackage{amsmath,amssymb,amsfonts}
\usepackage{mathtools}
\usepackage{algorithmic}
\usepackage{graphicx}
\usepackage{textcomp}
\usepackage{xcolor}
\usepackage{algorithm}
\usepackage{algorithmic}
\usepackage{color}
\newcommand{\re}[1]{{\color{black}{#1}}}
\newcommand{\ch}[1]{{\color{black}{#1}}}

\def\BibTeX{{\rm B\kern-.05em{\sc i\kern-.025em b}\kern-.08em
    T\kern-.1667em\lower.7ex\hbox{E}\kern-.125emX}}
\begin{document}

\title{Continual Horizontal Federated Learning for Heterogeneous Data\\
}

\author{\IEEEauthorblockN{Junki Mori}
\IEEEauthorblockA{\textit{Secure System Research Laboratories} \\
\textit{NEC Corporation}\\
Kanagawa, Japan \\
junki.mori@nec.com}
\and
\IEEEauthorblockN{Isamu Teranishi}
\IEEEauthorblockA{\textit{Secure System Research Laboratories} \\
\textit{NEC Corporation}\\
Kanagawa, Japan \\
teranisi@nec.com}
\and
\IEEEauthorblockN{Ryo Furukawa}
\IEEEauthorblockA{\textit{Secure System Research Laboratories} \\
\textit{NEC Corporation}\\
Kanagawa, Japan \\
rfurukawa@nec.com}
}


\maketitle

\begin{abstract}
Federated learning is a promising machine learning technique that enables multiple clients to collaboratively build a model without revealing the raw data to each other. Among various types of federated learning methods, horizontal federated learning (HFL) is the best-studied category and handles homogeneous feature spaces. However, in the case of heterogeneous feature spaces, HFL uses only common features and leaves client-specific features unutilized. In this paper, we propose a HFL method using neural networks named continual horizontal federated learning (CHFL), a continual learning approach to improve the performance of HFL by taking advantage of unique features of each client. CHFL splits the network into two columns corresponding to common features and unique features, respectively. It jointly trains the first column by using common features through vanilla HFL and locally trains the second column by using unique features and leveraging the knowledge of the first one via lateral connections without interfering with the federated training of it. We conduct experiments on various real world datasets and show that CHFL greatly outperforms vanilla HFL that only uses common features and local learning that uses all features that each client has.  

\end{abstract}

\begin{IEEEkeywords}
federated learning, continual learning, heterogeneous features
\end{IEEEkeywords}

\section{Introduction}
With the increasing concerns for data privacy in machine learning, \textit{federated learning} (FL) \cite{FL} has received a lot of attention as a privacy-preserving machine learning framework to enable multiple clients (e.g., hospitals) to collaboratively train a model without sharing raw data. Existing FL approaches mainly focus on the scenario where each client has homogeneous data, i.e., the data with the same feature space. This setting is called \textit{horizontal federated learning} (HFL) \cite{FLsurvey}. However, in many real world applications, the feature spaces of the data owned by different clients may not be completely the same. Their data may differ in the feature space except for some common features. Consider the following example \cite{HFTL}. A hospital may share common features of patients such as age, blood pressure, etc. with nursing homes and physical examination centers. However, each medical institution also holds additional unique features such as prescription and diet information, as well. \re{Applying existing HFL approaches to this setting is not efficient in terms of the performance because they can only use the common features and leaves the unique features unutilized. We need more efficient approaches to address this problem.}

\re{There are some methods to handle heterogeneous data, i.e., to utilize all features for each client such as \textit{vertical federated learning} (VFL) \cite{FLsurvey} and \textit{federated transfer learning} (FTL) \cite{FLsurvey, FTL}.} VFL assumes that each client has the same samples of data with unique feature space and jointly trains a model by fully utilizing distributed features of the overlapping samples. However, the overlapping samples are usually limited, and identifying their IDs among all clients can violate privacy problem in some cases \cite{intersection}. Moreover, in VFL, clients need to collaborate even during the inference phase because a VFL model takes all features as input. \re{From these perspectives, VFL connot be an efficient method to solve the problem we consider. On the other hand, FTL can handle all samples as well as all features and conduct inference locally, but most FTL methods need the overlapping samples as with VFL.}

Gao et al. \cite{HFTL} proposed a new technique, \textit{heterogeneous federated transfer learning} (HFTL), which addresses the heterogeneous feature space problem with feature co-occurrence only. In HFTL, each client mutually reconstructs the missing features by using exchanged feature mapping models. Then, using all features, the clients conduct HFL to obtain a label prediction model and merge the feature mapping model and the label prediction model into a final model that can be used locally during the inference phase. \re{However, in \cite{HFTL}, HFTL is applied only to a logistic regression model, and is difficult to apply to more complex models such as \textit{neural network} (NN)} because HFTL needs to apply multi-party computation \cite{MPC} to exchanging the feature mapping model.

In this paper, we propose \textit{continual horizontal federated learning} (CHFL), a NN-based HFL method to relax the same feature space assumption using \textit{continual learning} (CL) \cite{CLsurvey}. To solve the problem of different input features for different clients, we divide the features into common features and unique features for each client. Then, we jointly train a common network that takes common features as input by using vanilla HFL and locally train a unique network that takes unique features as input. We utilize an effective model architecture used in a CL method, \textit{progressive neural network} (PNN) \cite{PNN}, for training global and local networks, to combine each of their layer outputs and transfer the knowledge of the global network into each local network without obstructing the training of the global network. \re{In CHFL, clients do not need overlapping samples and can conduct inference locally using all of their own features.} Through extensive experiments using five real world datasets, we show that our method outperforms two baselines: a HFL approach based on common features only and a local learning approach.

\section{Related Work}
In this section, we focus on reviewing works that are closely related to our approach.

\subsection{Federated Learning (FL)}
FL is a distributed machine learning framework proposed by Google \cite{FL}, where multiple clients collaboratively train a model without sharing their data. On the basis of the distribution characteristics of the data, FL approaches can be categorized into three types as first introduced by Yang et al. \cite{FLsurvey}: horizontal federated learning, vertical federated learning, and federated transfer learning. 

\subsubsection{Horizontal Federated Learning (HFL)}
HFL assumes the homogeneous data scenario, which implies the feature spaces of data in different
clients are the same. If the data samples owned by different clients greatly differ, their collaborative training becomes advantageous to the machine learning model because of the increased number of the dataset samples. There are well-known HFL methods such as FedAvg \cite{FL} and FedProx \cite{fedprox}. In FedAvg, each client locally trains a deep learning model with their local data and uploads their trained model parameters to the server. The server updates the global model by averaging the parameters of the local models and sends it to all clients. This process is one iteration and is repeated until the convergence. Since the parameters shared with the server are only  the gradients, the data privacy is preserved.

FedProx is a FL method to address the problem of lacking an independent and identically distributed (IID) data assumption across clients, i.e., non-IID data problem. It adds the proximal term to the learning objective to reduce the potential weight divergence defined by Zhao et al. \cite{noniid}. To address non-IID setting, there are studies in different directions: \textit{personalized federated learning} \cite{mocha, perfedavg, pfedme, apfl, CFL, fedfomo}. These approaches aim to train optimal personalized models for each client, not a single model. From this perspective, our method that utilizes unique features of each client can be considered as a kind of personalized FL.  

\subsubsection{Vertical Federated Learning (VFL)}
VFL is in the heterogeneous data scenario where data owned by different clients have the same sample IDs but disjoint feature spaces. In the process of VFL, each client locally computes their model intermediate outputs that will be aggregated in a privacy-preserving way at a server or certain client to obtain the loss. Then, each client downloads the updated information obtained by minimizing the loss and updates their local model. In this way, a VFL model using all features is built. There are VFL studies to handle various models such as linear/logistic regression \cite{SLR, LR, LR2, LR3, LR4}, decision trees \cite{secureboost, tree, tree2, tree3, federboost}, neural networks \cite{fdml, NN, pyvertical, transnet}, and other non-linear models \cite{non-linear, fedv}. Chu and Zhang \cite{self-taught} and Kang et al. \cite{fedmvt} proposed VFL methods to utilize the data with non-overlapping sample IDs. Our research motivation is close to theirs because our method is a HFL method to utilize the uncommon features.

\subsubsection{Federated Transfer Learning (FTL)}
FTL applies to the scenario where two clients each have data that differ in not only sample ID space but also feature space. FTL transfers the knowledge of the source client to the target client who has only limited samples so as to be able to infer labels using their own features only. The first FTL method was proposed by by Liu et al. \cite{FTL}, where a common representation between the two feature spaces is learned using the overlapped samples. \ch{Gao et al. \cite{MPDL} proposed a dual learning framework to solve the heterogeneous feature space problem by missing data completion. Unlike our CHFL, these approaches \cite{FTL}, \cite{MPDL} need overrapped samples.} Gao et al. \cite{HFTL} considered the case where a part of feature space is only common among clients and proposed a method to map different feature spaces into the same ones before performing HFL. 

\subsection{Continual Learning (CL)}
Current deep learning algorithms perform excellently on a given single task, but learning new tasks continually is a hard problem because they tend to forget earlier tasks, which is called \textit{catastrophic forgetting} \cite{catas}. CL is a novel approach to address this problem by preserving the previously learned knowledge using various types of information of the past tasks. There are three types of CL methods based on how task-specific information is stored and used throughout the sequential learning process \cite{CLsurvey}: \re{\textit{replay methods} (storing and replaying previous task samples while learning a new task), \textit{regularization-based methods} (adding a regularization term in the loss function to consolidate previous knowledge), and \textit{parameter isolation methods} (dedicating different model parameters to each task).} PNN \cite{PNN} is a kind of parameter isolation methods.

\ch{There are several studies that combine CF and FL \cite{FedWeit, FCL2, FCL3, FCL4, FCL5}. In \cite{FedWeit, FCL2, FCL3}, the authors consider the scenario where each client has access to a private sequence of task. Researches in another direction use CL methods to solve weight divergence in FL due to the different data distribution problem \cite{FCL4, FCL5}. We use CL method for a different purpose, i.e., addressing the heterogeneous feature space problem.}

\section{Preliminaries}
\re{Throughout this paper, we denote the model parameters of a neural network with $L$ layers including the output layer as $\theta \coloneqq (W^{(1)}, \cdots,W^{(L)}, b^{(1)}, \cdots, b^{(L)})$, where $W^{(i)}$ and $b^{(i)}$ are the weight matrix and the bias vector of the $i$-th layer. Dataset is denoted as $(X, Y)$, where $X \in \mathbb{R}^{N \times d}$ is a set of $N$ data samples with the $d$-dimensional feature space and $Y \in \mathbb{R}^{N \times C}$ is a set of the corresponding one-hot encoding ground truth labels for $C$ classes. We use $F_{\theta}\colon \mathbb{R}^d\rightarrow \mathbb{R}^C$ to represent the whole network with parameters $\theta$.}
\re{\subsection{Federated Averaging (FedAvg)}}
\re{We introduce the most popular HFL method, FedAvg \cite{FL}. Consider $K$ clients $\{C_k\}_{k=1}^K$, each having dataset $(X_k, Y_k)$, where each $X_k$ has the same feature space. At iteration $t$ in FedAvg, the server sends a global model with parameters $\theta^{t}$ to all clients. Each client $C_k$ updates $\theta^{t}$ for given local epochs $E$ to obtain the local version of the global model $\theta_{k}^t$ by minimizing the cross entropy loss $\mathcal{L}(F_{\theta_{k}^{t}}(x_k),y_k)$ evaluated at mini-batch $(x_k, y_k)$ of $(X_k, Y_k)$. Then, it returns $\theta_{k}^t$ to the server that will aggregate each $\theta_{k}^t$ to obtain $\theta^{t+1}$ as
\begin{equation}
    \theta^{t+1} = \frac{1}{K}\sum_{k=1}^K \theta_{k}^t.
\end{equation}
This process is iterated $T$ times. Each client uses the common model $\theta^T$ for inference. Detailed algorithm is shown in Algorithm \ref{alg:FedAvg}. Here, we only consider the case where all clients always participate in model aggregation because in this paper the number of clients is set to at most ten.

\begin{algorithm}
\caption{FedAvg: Federated Averaging \cite{FL}}
\label{alg:FedAvg}
\begin{algorithmic}[1]
\REQUIRE Datasets $\{(X_k, Y_k)\}_{k=1}^K$ of $K$ clients $\{C_k\}_{k=1}^K$, parameters $\theta^0$, number of communication rounds $T$, number of local epochs $E$, learning rate $\eta$
\ENSURE \ 

Trained parameters $\theta^T$

\STATE \textbf{Server do:}
\FOR {$t=0, \cdots, T-1$}
\FOR {$k=1, \cdots, K$ \textbf{in parallel}}
\STATE Send $\theta^t$ to $C_k$
\STATE Do \textbf{ClientLocalLearning}($k$, $\theta^t$)
\ENDFOR
\STATE $\theta^{t+1} \leftarrow \frac{1}{K}\sum_{k=1}^K\theta_{k}^t$
\ENDFOR
\STATE Return $\theta^T$
\STATE \textbf{ClientLocalLearning}($k$, $\theta^t$):
\STATE $\theta_{k}^t \leftarrow \theta^t$
\FOR {epoch $e=1, \cdots, E$}
\FOR {each batch $(x_k, y_k)$ of $(X_k, Y_k)$}
\STATE $\theta_{k}^t \leftarrow \theta_{k}^t-\eta \nabla \mathcal{L}(F_{\theta_{k}^{t}}(x_k),y_k)$
\ENDFOR
\ENDFOR
\STATE Return $\theta_{k}^t$ to server

\end{algorithmic}
\end{algorithm}
}

\subsection{Progressive Neural Network (PNN)}
\label{sec:PNN}
We review PNN \cite{PNN}, a simple and powerful CL method belonging to parameter isolation methods that instantiates a new neural network for each task.

\begin{figure}[tbp]
\centerline{\includegraphics[width=0.85\linewidth]{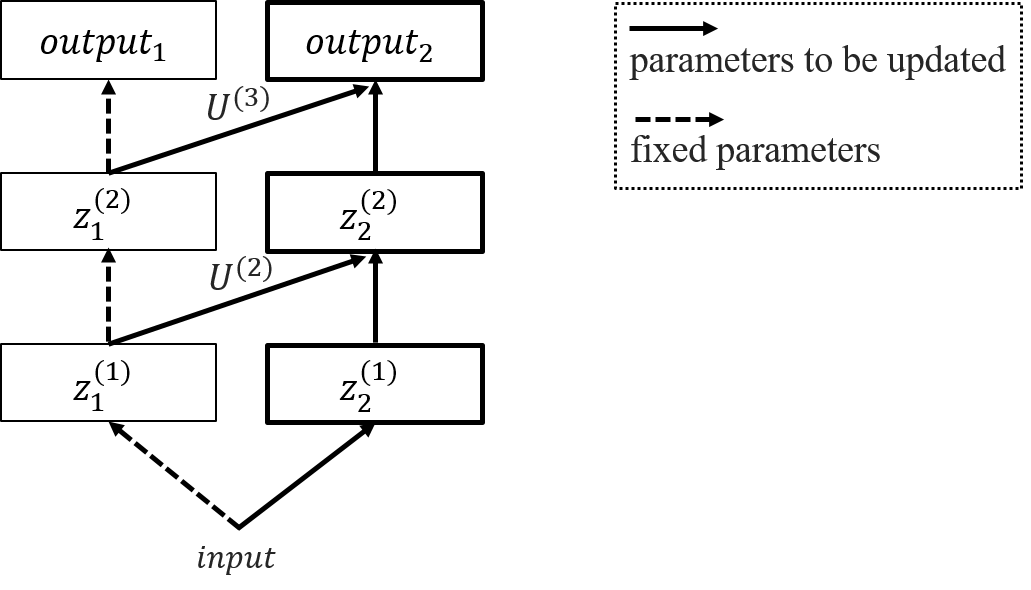}}
\caption{Description of a progressive neural network with three layers in the case of two tasks.}
\label{PNN}
\end{figure}

Formally, CL setting considers a sequence of tasks indexed by $s \in \{1, \cdots, S\}$. The learner receives training data of only one task at a time to train a model until the convergence. Data $(X_s, Y_s)$ for each task $s$ is sampled from different distributions, where each $X_s$ has the same feature space. Consider the case of $S=2$. PNN in this case is shown in Fig.~\ref{PNN}. PNN adds a new neural network (a \textit{column}) when learning a new task. We denote the parameters and the $i$-th layer's output of a column for task $s$ by $\theta_s$ and $z_s^{(i)} \in \mathbb{R}^{n_i}$, respectively, where $n_i$ is the number of units at the $i$-th layer. First, PNN trains $\theta_1$ using $(X_1, Y_1)$. Next, the parameters $\theta_1$ are fixed, and  $\theta_2$ is trained using $(X_2, Y_2)$, where the \mbox{$(i+1)$-th} layer receives input from both $z_2^{(i)}$ and $z_1^{(i)}$ via lateral connections. Therefore, $z_2^{(i+1)}$ for $1\leq i\leq L-1$ is computed as follows:
\begin{equation}
\label{output}
    z_2^{(i+1)} = a(W_2^{(i+1)}z_2^{(i)}+b_2^{(i+1)}+U^{(i+1)}z_1^{(i)}),
\end{equation}
where $a$ is an activation function, $W_s^{(i)} \in \mathbb{R}^{n_i \times n_{i-1}}$ and $b_s^{(i)} \in \mathbb{R}^{n_i}$ are the weight matrix and the bias vector of the $i$-th layer of $\theta_s$, respectively, and $U^{(i)} \in \mathbb{R}^{n_i \times n_{i-1}}$ is the \textit{lateral connection matrix} from the $(i-1)$-th layer of $\theta_1$ to the $i$-th layer of $\theta_2$. In this way, PNN transfers the knowledge of task $1$ to task $2$ without catastrophic forgetting.

\subsection{Problem Definition}
\label{problem}
We define the problem considered through this paper. Consider $K$ clients $\{C_k\}_{k=1}^K$, each having dataset $(X_k, Y_k)$. We denote the feature space, the label space, and sample ID space for client $C_k$ by $\mathcal{X}_k$, $\mathcal{Y}_k$, and $\mathcal{I}_k$, respectively. 

\begin{figure}[tbp]
\centerline{\includegraphics[width=\linewidth]{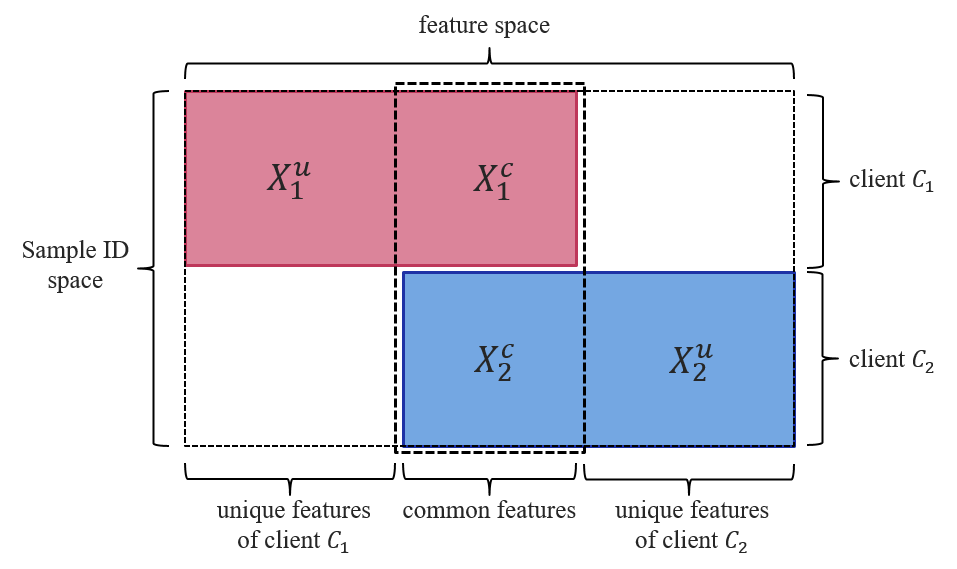}}
\caption{View of dataset structure in the two client scenario.}
\label{dataspace}
\end{figure}

We assume that the label spaces $\{\mathcal{Y}_k\}_k$ are the same and the feature spaces $\{\mathcal{X}_k\}_k$ are different but have common overlapping space:
\begin{equation}
    \mathcal{Y}_1 = \mathcal{Y}_2 = \cdots =\mathcal{Y}_K,
\end{equation}
\begin{equation}
    \bigcap_{k=1,\dots,K} \mathcal{X}_k \neq \emptyset.
\end{equation}
Let each $X_k$ be split into two parts, $X_k^c$ in the common feature space and $X_k^u$ in the unique feature space of client $C_k$. Here, we assume that all clients have already found or know the common feature space. Since the objective of HFL is to improve the performance of the machine learning model by increasing the number of data samples, we assume that each sample ID space has no intersection for simplicity:
\begin{equation}
    \mathcal{I}_k \cap{\mathcal{I}_l} = \emptyset, \ \forall k \neq l.
\end{equation}
Fig.~\ref{dataspace} shows the view of dataset structure in the case of two clients.

In this setting, vanilla HFL collaboratively trains a federated neural network using only the common features $\{X_k^c\}_k$. Hence, the unique features $\{X_k^u\}_k$ are not utilized. We aim to improve the performance of HFL model by utilizing not only common features but also unique features for each client. \re{Note that the method we seek has the following characteristics, unlike usual VFL and FTL:
\begin{itemize}
    \item Overlapping samples, i.e., samples with common IDs are not needed.
    \item Each client finally obtains a unique model and can conduct inference locally by using it and their own features only. 
\end{itemize}
}

\section{Proposed Approach}
We introduce the proposed approach, CHFL to address the heterogeneous data problem defined in Section \ref{problem}.

\subsection{Idea}
Data owned by each client differ in the feature space, \re{and therefore a vanilla HFL method cannot utilize all features.} To address this problem, we split the neural network for each client into two columns: \textit{unique column} and \textit{common column}, which take unique features and common features as inputs, respectively. We aim to jointly train the parameters of the common column by HFL and locally train the parameters of the unique column without interfering with the training of the parameters of the common column. We can consider this issue as learning a unique task (inferring with unique features) without forgetting the knowledge of a common task (inferring with common features). Inspired by this observation, we apply the CL method, especially PNN \cite{PNN} architecture, to this setting. \re{Utilizing PNN architecture enables a common column to be built collaboratively without being affected by the unique features of each client, and an unique column to leverage the federated knowledge of the common column via lateral connections, which leads to the CHFL being able to utilize both common features and unique features. In the following, we present more specific model architecture and detailed algorithm of CHFL.}

\begin{figure}[t]
\centerline{\includegraphics[width=0.95\linewidth]{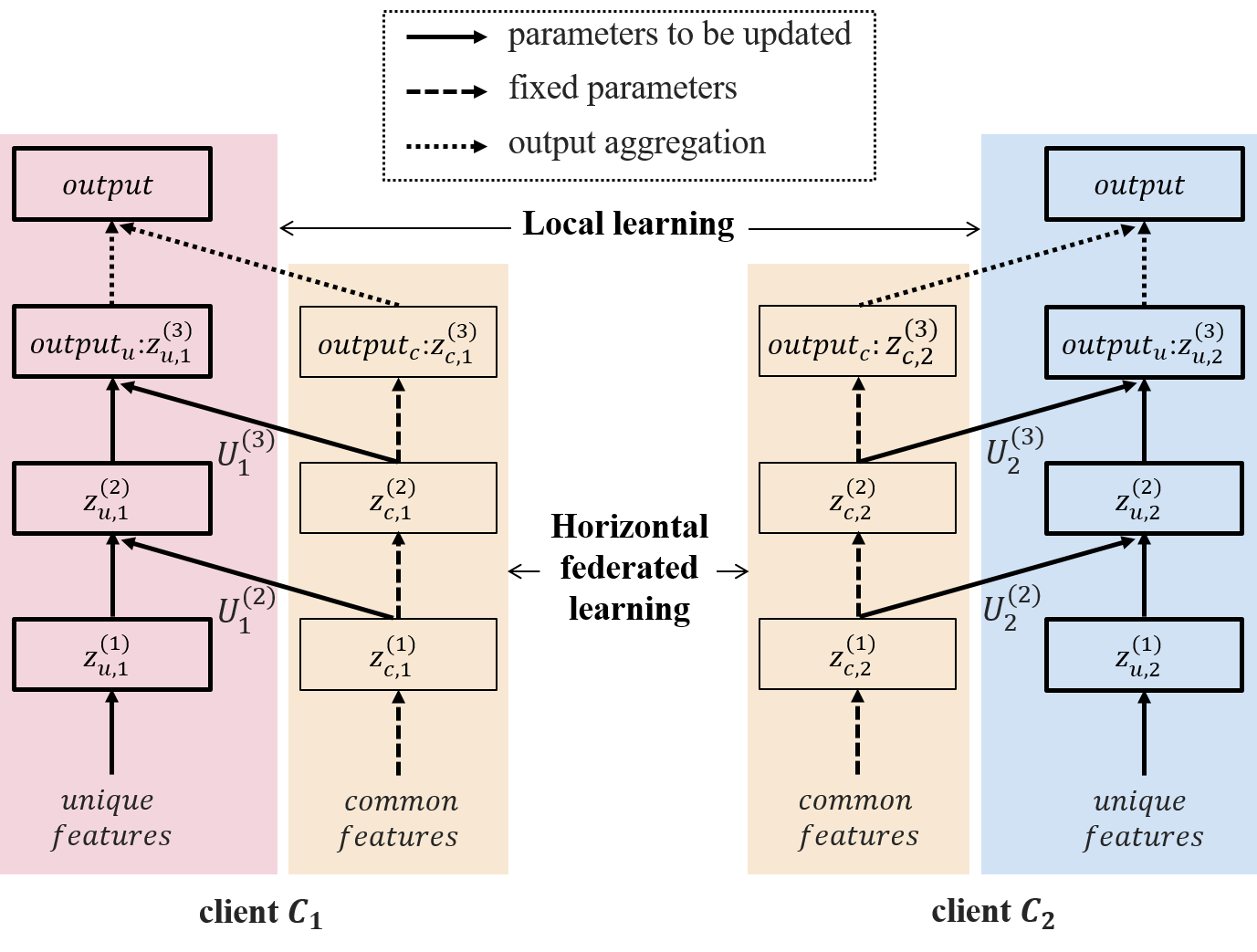}}
\caption{Architecture of CHFL in the case of $K=2$ and $L=3$.}
\label{proposed}
\end{figure}

\re{
\subsection{Model Architecture}
Fig. \ref{proposed} shows the architecture of CHFL. CHFL has three components for each client $C_k$: common column with parameters $\theta_c$, unique column with parameters $\theta_{u,k}$, and lateral connection matrices $U_k\coloneqq\{U_k^{(i)}\}_{i=2}^L$, where $L$ is the number of layers of $\theta_c$ and $\theta_{u,k}$. The unique column and lateral connection matrices are locally updated. We use notation $\Theta_k\coloneqq(\theta_c, \theta_{u,k}, U_k)$ as a whole set of parameters for client $C_k$. The $i$-th layer outputs of the common column and unique column for client $C_k$ are denoted as $z_{c,k}^{(i)}$ and $z_{u,k}^{(i)}$, respectively. 

As with PNN (see Section \ref{sec:PNN}), the lateral connection matrices are used to connect the intermediate outputs of common column and unique column. That is, $z_{u,k}^{(i+1)}$ is computed in the same way as (\ref{output}):
\begin{equation}
\label{output2}
    z_{u,k}^{(i+1)} = a\left(W_{u,k}^{(i+1)}z_{u,k}^{(i)}+b_{u,k}^{(i+1)}+U_k^{(i+1)}z_{c,k}^{(i)}\right).
\end{equation}
On the other hand, unlike PNN, each column takes different features as input. Moreover, each final output is aggregated because the objective of inference for each column is the same and therefore aggregation improves the prediction. Therefore, for input $(x_k^c, x_k^u)$ of client $C_k$, the final output of the whole network $\Theta_k$ is represented as follows:
\begin{equation}
\label{agg}
    F_{\Theta_k}(x_k^c, x_k^u) = \sigma \left(z_{c,k}^{(L)} + z_{u,k}^{(L)}\right),
\end{equation}
where $\sigma$ is a softmax function.
}

\subsection{Algorithm}
\label{sec:alg}
\re{We review the algorithm of CHFL in detail. In CHFL based on FedAvg, each client $C_k$ collaboratively trains common column $\theta_c$ using $(X_k^c, Y_k)$ according to FedAvg defined in Algorithm \ref{alg:FedAvg}. In parallel to the update of $\theta_c$, each client locally trains their unique column $\theta_{u,k}$ and lateral connection matrices $U_k$. Algorithm \ref{alg:CHFL} shows this process in detail. At iteration $t$, the server first sends the current parameters of common column $\theta_c^t$ to all clients as with FedAvg. After each client updates the local version of the common column $\theta_{c,k}^t$ at each mini-batch $(x_k^u, x_k^c, y_k)$ of $(X_k^u, X_k^c, Y_k)$, it computes the output of the whole network $F_{\Theta_{k}^{t}}(x_k^c, x_k^u)$ according to (\ref{agg}), where $\Theta_k^t\coloneqq(\theta_c^t, \theta_{u,k}, U_k)$. Then, each client fixes the parameters $\theta_{c,k}^t$ and updates $\theta_{u,k}$ and $U_k$ by minimizing the loss $\mathcal{L}(F_{\Theta_{k}^{t}}(x_k^c, x_k^u),y_k)$. The server receives $\theta_{c,k}^t$ from each client $C_k$ and aggregates them to obtain $\theta_{c}^{t+1}$. After $T$ iterations are ended, each client uses $F_{\Theta_{k}^{t}}$ for the inference model using both common features and unique features.}

Here, we introduce a parameter $0\leq\mu\leq 1$ to control the strength of the connection between the common column and unique column. If there is no correlation between common features and unique features, the training of the unique column may be interfered with by lateral connections. Therefore, we replace the lateral matrices $U_k$ by $\mu U^t$, where $\mu$ is not a learning parameter. If $\mu$ equals $0$, common column and unique column are connected only through the final output aggregation (\ref{agg}) as with FDML \cite{fdml}, a VFL method using neural networks \re{although there is the difference that FDML aggregates the outputs across clients}.  

\re{
\begin{algorithm}
\caption{CHFL: Continual horizontal federated learning}
\label{alg:CHFL}
\begin{algorithmic}[1]
\REQUIRE Datasets $\{(X_k^u, X_k^c, Y_k)\}_{k=1}^K$ of $K$ clients $\{C_k\}_{k=1}^K$, parameters of common column $\theta_c^0$, parameters of unique columns of $K$ clients $\{\theta_{u,k}\}_{k=1}^K$, lateral connection matrices of $K$ clients $\{U_k\}_{k=1}^K$, number of communication rounds $T$, number of local epochs $E$, learning rate $\eta$
\ENSURE \ 

Trained parameters $\theta_c^T$ and $\{\theta_{u,k}\}_{k=1}^K$,

Trained matrices $\{U_k^T\}_{k=1}^K$
\STATE \textbf{Server do:}
\FOR {$t=0, \cdots, T-1$}
\FOR {$k=1, \cdots, K$ \textbf{in parallel}}
\STATE Send $\theta_c^t$ to $C_k$
\STATE Do \textbf{ClientLocalLearning}($k$, $\theta_c^t$)
\ENDFOR
\STATE $\theta_c^{t+1} \leftarrow \frac{1}{K}\sum_{k=1}^K\theta_{c,k}^t$
\ENDFOR
\STATE Return $\theta_c^T$
\STATE \textbf{ClientLocalLearning}($k$, $\theta_c^t$):
\STATE $\theta_{c,k}^t \leftarrow \theta_c^t$
\FOR {epoch $e=1, \cdots, E$}
\FOR {each batch $(x_k^u, x_k^c, y_k)$ of $(X_k^u, X_k^c, Y_k)$}
\STATE $\theta_{c,k}^t \leftarrow \theta_{c,k}^t-\eta \nabla \mathcal{L}(F_{\theta_{c,k}^{t}}(x_k^c),y_k)$
\STATE $\theta_{u,k} \leftarrow \theta_{u,k}-\eta \nabla_{\theta_{u,k}} \mathcal{L}(F_{\Theta_{k}^{t}}(x_k^c, x_k^u),y_k)$
\STATE $U_{k} \leftarrow U_{k}-\eta \nabla_{U_{k}} \mathcal{L}(F_{\Theta_{k}^{t}}(x_k^c, x_k^u),y_k)$
\ENDFOR
\ENDFOR
\STATE Return $\theta_{c,k}^t$ to server

\end{algorithmic}
\end{algorithm}
}

\section{Experiments}
In this section, we evaluate the empirical performance of our CHFL in five datasets and show that it outperforms two baseline methods: a vanilla HFL using only common features and a local learning using both the common features and unique features. We first introduce experimental settings including datasets, compared methods, and implementation information. 

\subsection{Experimental Settings}
\subsubsection{Datasets}
We conducted experiments on five public datasets: (i) forest covertype dataset \cite{covertype}, (ii) sensorless drive diagnosis dataset \cite{uci}, (iii) aloi \cite{aloi}, (iv) helena \cite{openml}, and (v) jannis \cite{openml}. We use the training dataset of covertype from Kaggle as a whole dataset so that the number of data samples for each class is the same. Detailed information of these datasets is shown in Table \ref{dataset}.

\begin{table}[tbp]
\caption{Types of datasets}
\label{dataset}
\begin{center}
\begin{tabular}{|c|c|c|c|}
\hline
\textbf{Dataset} & \textbf{Sample Num.}& \textbf{Feature Num.} & \textbf{Class Num.} \\
\hline
\hline
covertype & 15,121 & 54 & 7 \\
\hline
sensorless & 58,509 & 48 & 11 \\
\hline
aloi & 108,000 & 128 & 1000 \\
\hline
helena & 65,196 & 27 & 100 \\
\hline
jannis & 83,733 & 54 & 4 \\
\hline
\end{tabular}
\end{center}
\end{table}

\begin{table*}[!ht]
\caption{Performance of each method in five datasets}
\label{main_result}
\begin{center}
\begin{tabular}{|c|c|c|c|c|}
\hline 
\textbf{Dataset} & \multicolumn{2}{|c|}{\textbf{Baseline}}& \multicolumn{2}{|c|}{\textbf{CHFL}} \\
\cline{2-5}
& Common & Local & $\mu=0$ & $\mu > 0$ \\
\hline
\hline
covertype &\ 0.6218 $\pm$ 0.1086  \ &\  0.6407 $\pm$ 0.0594 \ &\  0.6744 $\pm$ 0.0731 \ &\  \textbf{0.6843} $\pm$ 0.0661 \ \\
\hline
sensorless &\  0.9762 $\pm$ 0.0024 \ &\  0.9526 $\pm$ 0.0046 \ &\  0.9855 $\pm$ 0.0017 \ &\  \textbf{0.9863} $\pm$ 0.0020 \ \\
\hline
aloi &\  0.7352 $\pm$ 0.0366 \ &\  0.7093 $\pm$ 0.0165 \ &\  0.7927 $\pm$ 0.0196 \ &\  \textbf{0.7954} $\pm$ 0.0197 \ \\
\hline
helena &\  0.2604 $\pm$ 0.0147 \ &\  0.2732 $\pm$ 0.0100 \ &\  0.2868 $\pm$ 0.0086 \ &\  \textbf{0.2902} $\pm$ 0.0088 \ \\
\hline
jannis &\  0.6449 $\pm$ 0.0184 \ &\  0.6341 $\pm$ 0.0090 \ &\  0.6551 $\pm$ 0.0128 \ &\  \textbf{0.6572} $\pm$ 0.0111 \ \\
\hline
\end{tabular}
\end{center}
\end{table*}

\begin{table*}[!ht]
\caption{Performance of each method with different number of clients}
\label{clients_result}
\begin{center}
\begin{tabular}{|c|c|c|c|c|}
\hline 
\textbf{Clients} & \multicolumn{2}{|c|}{\textbf{Baseline}}& \multicolumn{2}{|c|}{\textbf{CHFL}} \\
\cline{2-5}
& Common & Local & $\mu=0$ & $\mu > 0$ \\
\hline
\hline
2 & \ 0.6420 $\pm$ 0.1098 \ &\  0.7649 $\pm$ 0.0144 \ &\  0.7475 $\pm$ 0.0356 \ &\  \textbf{0.7748} $\pm$ 0.0224 \ \\
\hline
5 &\ 0.6218 $\pm$ 0.1086 \ &\  0.6407 $\pm$ 0.0594 \ &\  0.6744 $\pm$ 0.0731 \ &\  \textbf{0.6843} $\pm$ 0.0661 \ \\
\hline
10 &\  0.6014 $\pm$ 0.1112 \ &\  0.5628 $\pm$ 0.0854 \ &\  0.6290 $\pm$ 0.0921 \ &\  \textbf{0.6291} $\pm$ 0.0914 \ \\
\hline
\end{tabular}
\end{center}
\end{table*}

\begin{table*}[!ht]
\caption{Performance of each method with different correlations}
\label{correlation_result}
\begin{center}
\begin{tabular}{|c|c|c|c|c|}
\hline 
\textbf{Correlation} & \multicolumn{2}{|c|}{\textbf{Baseline}}& \multicolumn{2}{|c|}{\textbf{CHFL}} \\
\cline{2-5}
& Common & Local & $\mu=0$ & $\mu > 0$ \\
\hline
\hline
Maximum & \ 0.2717 $\pm$ 0.0062 \ &\  0.2811 $\pm$ 0.034 \ &\  0.2867 $\pm$ 0.0047 \ &\  \textbf{0.2916} $\pm$ 0.0039 \ \\
\hline
Median &\  0.2797 $\pm$ 0.0168 \ &\  0.2926 $\pm$ 0.0080 \ &\  0.2999 $\pm$ 0.0104 \ &\  \textbf{0.3049} $\pm$ 0.0091 \ \\
\hline
Minimum &\  0.1793 $\pm$ 0.0133 \ &\  0.2476 $\pm$ 0.0019 \ &\  0.2695 $\pm$ 0.0016 \ &\  \textbf{0.2710} $\pm$ 0.0013 \ \\
\hline
\end{tabular}
\end{center}
\end{table*}

\subsubsection{Compared Methods}
We compare CHFL with following two baseline methods:
\begin{itemize}
    \item Common: A vanilla HFL method over common features among different clients is used. We use FedAvg as a vanilla HFL.
    \item Local: Each client locally trains a neural network using all of their own features, i.e., both common features and unique features. 
\end{itemize}
In all methods including CHFL, we use simple fully-connected neural networks composed of three hidden layers with 512, 256, and 128 neurons, respectively. In CHFL, we use the same architecture with the above mentioned size for the common column and unique column.

\subsubsection{Implementations}
All datasets are split randomly with 60\%, 20\%, and 20\%
for training, validation, and testing, respectively. We split and distribute training and validation datasets evenly to each client. Each client tunes hyperparameters such as learning rate and $\mu$ of CHFL using their own validation dataset. We evaluate the performance of each method by average testing accuracy of all clients. \re{In all experiments, we randomly split features into common features and others. The ratio of the common features is set to 30\% unless explicitly specified.} The remaining features are divided into the number of clients, and each part is used as unique features of the corresponding client. We conduct experiments three times changing the way that features are split randomly and compute their average accuracy. Moreover, for each splitting, five experiments are performed and averaged. Throughout the experiments, the number of clients, the number of local epochs, and batch size are set to $5$, $5$, and $64$ by default, respectively . We use Adam as the optimizer. All experiments are conducted using the PyTorch 1.8 framework on a Tesla V100 GPU with 32-GB memory.

\subsection{Experimental Results}

\subsubsection{Main results}
Table \ref{main_result} shows the average test accuracy of five clients with standard deviations on five datasets. The highest average test accuracy is marked in bold. In CHFL, we compare two cases, $\mu=0$ and $\mu>0$ to investigate the effectiveness of lateral connections between two columns. In the case of $\mu>0$, we use $\mu$ to demonstrate the best performance on validation dataset. From Table \ref{main_result}, we can see that CHFL outperforms the two baselines for every dataset. Moreover, it is shown that non-zero $\mu$ improves CHFL, which means lateral connections effectively transfer the knowledge of the common column to the unique column. However, we can also see that even the case without lateral connections (i.e., the case of $\mu=0$) sufficiently improves baseline methods.

\subsubsection{Ratio of common features}
\begin{figure}[t]
\centerline{\includegraphics[width=0.9\linewidth]{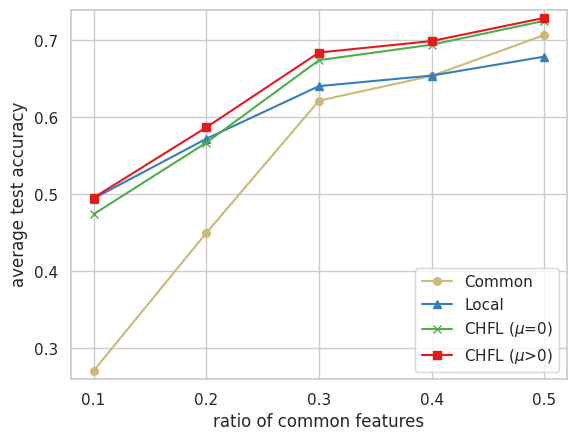}}
\caption{Variation of average test accuracy for covertype with respect to the ratio of the common features.}
\label{ratio}
\end{figure}

The effect of the ratio of common features on test accuracy for covertype is shown in Fig. \ref{ratio}. In Fig. \ref{ratio}, the ratio is changed from 10\% to 50\%. We can see that CHFL in the case of $\mu>0$ is always the best, but approaches baselines when the ratio of common features is small or large. This is because when the ratio of common features is small, the effect of HFL is small, and when the ratio of common features is large, the effect of using unique features is small. In fact, the test accuracy of Common is greatly improved as the ratio of common features increases. \re{We can also see that as the ratio of the common features increases, the difference in performance between CHFL ($\mu=0$) and CHFL ($\mu>0$) becomes smaller, i.e., the effect of lateral connections becomes smaller.}

\re{
\subsubsection{Number of clients}
We investigated the effect of the number of clients. The results in the case of 2, 5, 10 clients for covertype are shown in Table \ref{clients_result}. The ratio of common features is fixed to 30\%.
The results show that the difference of CHFL and baselines is small in the case of two clients and ten clients. The reason for this is similar to the experiment in Fig. \ref{ratio}: when there are many clients (few unique features), the effect of using unique features is small, and when there are few clients, the effect of HFL is small.}

\re{
\subsubsection{Strength of correlation}
As mentioned in Section \ref{sec:alg}, the strength of the correlation between the common features and unique features is considered to be related to the performance of CHFL. To confirm this intuition, we calculated the correlation coefficients between the common features and unique features using helena. First, we split features into common features and others. Then, we calculate the correlation coefficients between any column of the common features and any column of the remaining features and sum up them. This summation of the correlation coefficients was computed in every way of feature splitting. We conducted experiments in the three ways of feature splitting where the summation of the correlation coefficients is maximum, minimum, and median, respectively. The results are shown in Table \ref{correlation_result}. Comparing CHFL ($\mu=0$) and CHFL ($\mu>0$), we can see that there is little difference between them in the case of Minimum. Therefore, when the correlation between common features and unique features is weak, a simple method CHFL without lateral connections is enough.
}

\subsubsection{Comparison to another simple method}
In addition to the simple method, CHFL without lateral connections, we compared CHFL with lateral connections to another simple method. We consider the following method. As with CHFL, clients collaboratively train a common neural network using common features by applying FedAvg. In parallel to that, each client trains a unique neural network that takes the concatenation of the unique features and the output of the common network as input.
Thus, this method can also utilize all features. We evaluated the performance of this simple method for covertype and jannis in the same setting as Table \ref{main_result}, which resulted in the test accuracy of $0.6637 \pm{ 0.0709}$ and $0.6528 \pm{ 0.0111}$ , respectively. Comparing with Table \ref{main_result}, we find that this method is less effective than even CHFL without lateral connections.  Hence, we can see that our lateral connections and output aggregation are effective.

\section{Conclusions}
We proposed CHFL, a continual learning approach to address the heterogeneous data problem in HFL. CHFL prepares two neural networks corresponding to common features and unique features, respectively and laterally connect them while training the common network with HFL and locally training the unique network. Through the experiments on various datasets,  CHFL was shown to improve vanilla HFL method that only uses common features and outperform local learning that uses all features of the corresponding client.

\section*{Acknowledgment}
We are grateful to Tomoyuki Yoshiyama for useful discussions. We would also like to thank Batnyam Enkhtaivan and Kunihiro Ito for helpful comments on the manuscript.

\bibliographystyle{IEEEtran}
\bibliography{IEEEexample}

\end{document}